Bennett, CC, Doub TW, Bragg AD, et al. Data Mining Session-Based Patient Reported Outcomes (PROs) in a Mental Health Setting: Toward Data-Driven Clinical Decision Support and Personalized Treatment. *IEEE Health Informatics and Systems Biology Conference*. 2011.
http://ieeexplore.ieee.org/xpl/freeabs_all.jsp?arnumber=6061404&abstractAccess=no&userType=

# Data Mining Session-Based Patient Reported Outcomes (PROs) in a Mental Health Setting: Toward Data-Driven Clinical Decision Support and Personalized Treatment

Casey Bennett, M.A.[1,2], Thomas Doub, Ph.D[1], April Bragg, Ph.D.[1], Jason Luellen, Ph.D[1], Christina Van Regenmorter, M.S.W.[1], Jennifer Lockman, M.S.[1], Randall Reiserer Ph.D[1]

[1]Centerstone Research Institute
Nashville, TN, USA
Casey.Bennett@CenterstoneResearch.org

[2]School of Informatics and Computing
Indiana University
Bloomington, IN, USA

*Abstract*— The CDOI outcome measure – a patient-reported outcome (PRO) instrument utilizing direct client feedback – was implemented in a large, real-world behavioral healthcare setting in order to evaluate previous findings from smaller controlled studies. PROs provide an alternative window into treatment effectiveness based on client perception and facilitate detection of problems/symptoms for which there is no discernible measure (e.g. pain). The principal focus of the study was to evaluate the utility of the CDOI for predictive modeling of outcomes in a live clinical setting. Implementation factors were also addressed within the framework of the Theory of Planned Behavior by linking adoption rates to implementation practices and clinician perceptions. The results showed that the CDOI does contain significant capacity to predict outcome delta over time based on baseline and early change scores in a large, real-world clinical setting, as suggested in previous research. The implementation analysis revealed a number of critical factors affecting successful implementation and adoption of the CDOI outcome measure, though there was a notable disconnect between clinician intentions and actual behavior. Most importantly, the predictive capacity of the CDOI underscores the utility of direct client feedback measures such as PROs and their potential use as the basis for next generation clinical decision support tools and personalized treatment approaches.

## I. Introduction

Recent years have seen the proliferation of the concepts of patient-reported outcomes (PROs) and client feedback throughout the healthcare and behavioral healthcare spheres. Research has begun to illuminate their utility on the effectiveness of clinical treatment across multiple disorders and diseases, including depression, schizophrenia, diabetes, cancer, and speech problems, among others [1-5]. The basic concept is that utilizing outcomes and/or feedback directly from the patient, rather than filtered through the perspective of the clinician, can provide meaningful information around a patient's current status, progress, and prognosis.

More specifically, PROs have been suggested to benefit clinical practice via 1) facilitating the detection of overlooked problems, including some symptoms for which there is no discernible measure (e.g. pain), 2) providing an alternative window into treatment effectiveness that still utilizes standardized measures, 3) potentially improving patient-clinician communication, and 4) being unaffected by inter-rater reliability [1,6]. On the other hand, PROs have been criticized for 1) being too lengthy and burdensome on clinical workflow, and 2) the skepticism surrounding their clinical meaning and relevance [6]. Beyond this, even if PROs have no direct negative effects, their use may preclude the use of other, more meaningful measures given the limited clinical time available for outcomes collection [7].

CDOI (Client-Directed Outcome Informed) is a PRO developed by Miller et al. [8]. It was developed with the specific purpose of providing a valid yet brief outcome measurement system that could fit practically into real-world clinical practice. The CDOI is a session-based measure, collected electronically at every visit. It is comprised of two ultra-brief scales comprising 4 questions each – ORS (a measure of clinical symptomology and functioning) and SRS (a measure of therapeutic alliance). Previous research has shown ORS to compare favorably with other, longer outcome measures such as OQ45 [9,10], as well as the SRS comparing favorably to other, longer measures of therapeutic alliance such as the HAQ-II [11]. In concert with findings from other PRO studies, evaluation of the CDOI has shown that it can improve clinical outcomes [8,12].

The principal focus of the study was on evaluating the utility of the CDOI for predictive modeling of outcomes in a live clinical setting. The question exists what utility such an approach may hold with respect to clinical decision support (CDSS) recommendations, particularly a data-driven approach to CDSS that relies on data from live clinical systems and artificial intelligence algorithms that "learn" over time. Electronic health records themselves are only a first step. Techniques such as predictive modeling and data mining can detect patterns in the data that can be applied to new clients when they walk in the door, for example providing recommendations to clinicians about the treatment options most likely to be effective for a specific individual. This is, in essence, an individualized form of practice-based evidence. Another term for this is "personalized medicine" [13,14]. PROs and client feedback can play an important role in this sort of individualized approach [5].

Prior work in this area has primarily addressed the utility of genetic data to inform individualized care. However, it is likely that the next decade will see the integration of multiple sources of data – genetic, clinical, socio-demographic – to build a more complete profile of the individual, their inherited risks, and the environmental/behavioral factors associated with disorder and the effective treatment thereof [15]. Indeed, we already see the trend of combining clinical and genetic indicators in prediction of cancer prognosis as a way of developing cheaper, more effective prognostic tools [16-18].

PROs and client feedback fit into this larger paradigm around personalized medicine and clinical decision support technology (CDSS). However, evidence shows only a small minority of clinicians report collecting outcomes in any form [19]. Bickman [20] addressed these issues, noting that "currently everyone but the client appears to benefit from not having a measurement feedback system" (pp.1115). States and funders can claim they are providing for the needs of their citizens without worrying about the effectiveness (or ineffectiveness) of the resources spent. Provider organizations can justify their use of funds by claiming they use "evidence-based" practices. Clinicians and supervisors can evaluate their performance based on their own perceptions rather than actual data [20]. Establishing the utility of PROs and client feedback for making actual treatment decisions is critical to addressing such realities.

A secondary focus of the study was to examine the role of implementation factors within the framework of the Theory of Planned Behavior (TPB) [21] by linking adoption rates to implementation practices and clinician perceptions. This is a critical topic in understanding utilization of technology, research innovations, outcomes assessments (including PROs such as CDOI), and the like in real world settings [22-25]. Good technology that goes unused is of questionable success. There is still a need for better understanding of multi-pronged, flexible yet replicable implementation strategies and barriers thereof [26,27].

## II. METHODS

### A. Setting and Data

The CDOI was implemented at Centerstone, the largest community-based (outpatient clinic) mental healthcare provider in the United States seeing over 75,000 distinct individuals a year across 130 clinic locations in Indiana and Tennessee. Centerstone Research Institute is an arm of Centerstone devoted to integrating evidence and practice, conducting clinical research, developing clinical decision support tools, and building new healthcare informatics technologies, among other goals. Centerstone has a fully functional electronic health record (EHR) that maintains virtually all relevant patient records. Its clinical services operate under a mixture of both fee-for-service and case rate payment methodologies, including Medicare, Safety Net (Tennessee-sponsored health insurance for the population of seriously mentally ill who are ineligible for Medicaid), three distinct Medicaid payers (due to state subcontracting) with different sets of rules, a variety of Commercial payers, as well as an assortment of "other" payers such as county subsidies, DCS (Tennessee Dept. of Children Services), federal probation funds, and grants.

Approximately 150 clinicians were selected for inclusion in a 6-month pilot study from March 2010 thru August 2010. These included clinicians both in adult outpatient therapy as well as intensive in-home case management for children. For purposes of this analysis, data was limited to that collected for adults and adolescents over the age of 13 who received at least some therapy. The CDOI was integrated into the EHR and collected on a per session basis. As often occurs in real clinical practice, the CDOI was not always collected with fidelity at every single time point, and patients had the option to skip the ORS or SRS at any point if they so chose. The ORS was collected at the beginning of each session via a slider bar (scale 0-10) directly on the screen that the client could manipulate, in relation to each of the four questions that comprise the ORS scale [8]. The SRS was similarly collected, but at the end of each session via a slider bar (scale 0-10) directly on the screen that the client could manipulate, in relation to the four questions that comprise the SRS scale [8]. Furthermore, clients were identified as either "new" or "old" based on whether they had been seen within 90 days prior of the baseline CDOI at any Centerstone treatment facility or program.

For purposes of the data mining analysis, data was limited to those patients who received both a baseline CDOI and 3rd visit CDOI (based on previous "early change" findings, see the Results section), as well as having an additional CDOI at some point between the 5th visit and 10th visit (defined as the "final" CDOI). As such, all patients included in the sample were essentially examined based on their first 10 visits, which generally equates to a 3-4 month time period. The final sample was n=714 (n=253 for new patients only). On average, they received 7.2 visits during the 6 month pilot, with an average length of stay of 115 days (defined as baseline to final; patients may have continued to receive services outside the scope of the CDOI analysis).

### B. Data Analysis

Data was pulled nightly from the electronic health record and loaded into a data warehouse (DW) specifically for answering research and analytical questions. The DW also served as the basis for reporting and clinical feedback as specified in the implementation section below [14]. Subsequently, data was loaded into KNIME (Version 2.3.1) [28], an advanced data mining, modeling, and statistical software package. Data mining typically follows a standard process flow that can be broken into a number of main steps: data preparation, feature selection, model construction, and model evaluation. It should be noted that not all of these steps are performed all the time – for instance one may build models without any feature selection in order to evaluate the effect of feature selection on a particular dataset. Below, these steps are briefly outlined in the context of the current study; a more comprehensive overview of specific data mining strategies and methodologies can be found in any of a number of resources on the subject [29,30].

The primary analysis focused on clinical outcomes as measured by the change in ORS scores over time (research questions #1 and #2). The primary question was whether clients would obtain average or better outcome delta (change over time) from baseline to final visit (or vice versa, worse outcome delta). As such, the target variable (ORS delta final) was discretized into a binary variable of plus/minus the mean (equivalent to an equal bins classification approach). The consequences and assumptions of reduction to a binary classification problem are addressed in Boulesteix et al. [18], noting that the issues of making such assumptions are roughly equivalent to making such assumptions around normal distributions. All predictor variables (shown in Table I) were z-score normalized. Subsequently, all predictor variables were either 1) not discretized (labeled "Bin Target"), or 2) discretized via CAIM (Class-Attribute Interdependence Maximization). CAIM is a form of entropy-based discretization that attempts to maximize the available "information" in the dataset by delineating categories in the predictor variables that relate to classes of the target variable. By identifying and using patterns in the data itself, CAIM has been shown to improve classifier performance [31]. It should be noted that not all models are capable of handling both discretized and continuous variables, and thus both methods were not applied to all modeling methods. Additionally, some methods, such as certain kinds of neural networks or decision trees, may dynamically convert numeric variables into binary or categorical variables as part of their modeling process. As such, even when no pre-discretization was performed, it may have occurred within the modeling process itself. It should also be noted that practically all clients during the pilot received therapy services, which effectively excised all the q_therapy variables, including specific individual and group ones, from the analysis due to lack of variance.

TABLE I. VARIABLE LIST

| Variable | Description |
|---|---|
| bl_ors | Baseline ORS |
| bl_srs | Baseline SRS |
| third_delta_ors | ORS delta from baseline to third visit |
| third_delta_srs | SRS delta from baseline to third visit |
| gender | Gender of client |
| diag_cat | Primary Diagnostic Category (e.g. Mood, Anxiety, Substance Abuse) |
| age | Age of client |
| payor_grp | Payor Group (e.g. Medicaid, Medicare, Commercial, etc.) |
| county | County of primary clinic location of client |
| region_type | Urban or Rural |
| q_case_mgmt_bin | Binary (1/0) if client received any Case Mgmt services |
| q_medical_bin | Binary (1/0) if client received any Medical services |
| q_therapy_bin | Binary (1/0) if client received any Therapy services |
| q_ind_therapy_bin | Binary (1/0) if client received any Individual Therapy services |
| q_grp_therapy_bin | Binary (1/0) if client received any Group Therapy services |
| state | State of primary clinic location of client |
| final_delta_ors | Target (Dependent) variable, ORS delta from baseline to final |

Multiple models were constructed on the dataset to determine optimal performance using both native, built-in KNIME models as well as models incorporated from WEKA (Waikato Environment for Knowledge Analysis; Version 3.5.6) [30]. Models were generally run using default parameters, though some experimentation was performed. Of note, decay was set to true for MP Neural Networks, max_parents was set to 3 for Bayesian Network-K2, and number of nearest neighbors was set to 3 for K-Nearest Neighbors. Models tested included Naïve Bayes [30], HNB (Hidden Naïve Bayes) [32], AODE (Aggregating One-Dependence Estimators) [33], Bayesian Networks[30], Multi-layer Perceptron neural networks [30], Random Forests [34], J48 Decision Trees (a variant of the classic C4.5 algorithm) [35], Log Regression, and K-Nearest Neighbors [36]. Additionally, ensembles were built using a combination of Naïve Bayes, Multi-layer Perceptron neural network, Random Forests, K-nearest neighbors, and logistic regression, employing forward selection optimized by AUC (area under the curve) [37]. Voting by committee was also performed with those same five methods, based on maximum probability [38]. Voting by committee is a "meta-modeling" technique (like ensemble) that "combines" multiple models by allowing them to "vote" for the winning classification. It seeks to take advantage of the strengths of different modeling approaches while minimizing their drawbacks. Due to the number of models used, detailed explanations of individual methods are not provided here for brevity, but can be found elsewhere [29,30].

The last step was to evaluate model performance to rule out the possibility that statistical findings may be an artifact of capitalization on chance, which was performed using 10-fold cross-validation [30]. All models were evaluated using multiple performance metrics, including raw predictive accuracy; variables related to standard ROC (receiver operating characteristic) analysis such as AUC (area under the curve) and true/false positive rates [39]; and Hand's H [40]. The data mining methodology and reporting is in keeping with recommended guidelines [41,42], such as the proper construction of cross-validation, incorporation of feature selection within cross-validation folds, testing of multiple methods, and reporting of multiple metrics of performance, among others.

Additionally, some of the better performing models were evaluated using feature selection prior to modeling (but within each cross-validation fold). Feature selection is a key component in filtering out noisy and/or redundant variables from datasets and building parsimonious, explanatory models that retain generalizability. Particularly of interest was how clinical and demographic variables would compare in utility versus baseline outcome measures and "early change" delta (baseline to third visit) for predicting final outcome delta. The feature selection methods used include univariate filter methods (Chi-squared, Relief-F), multivariate subset methods (Consistency-Based –Best First Search, Symmetrical Uncertainty Correlation-Based Subset Evaluator) and wrapper-based (Rank Search employing Chi-squared and Gain Ratio). The advantages and disadvantages of these different types of feature selection are well-addressed elsewhere [43]. For purposes of this study, feature selection was only performed using a Naïve Bayes wrapper (where variables to be included in the final model were chosen by first building preliminary Naïve Bayes models using different selections of variables and selecting the feature set with the highest cross-validated performance) [44].

### C. Implementation Analysis

A clinician survey based on TPB was sent to all clinicians, with about half (n=66) responding. The survey included questions to assess the three domains of TPB as well as questions on intention to use the CDOI and years of experience as a practicing clinician (see Table II), each scored as a 4-point Likert scale. These data were then combined with other EHR data such as clinician age and gender. Adoption rates for each clinician (indicating the percent of the time the CDOI was completed when it should have been, not counting client opt-out skips) were calculated for November, 3 months after the end of the pilot and implementation efforts. Additionally, average outcome scores (including baseline ORS and SRS and ORS delta from baseline to final) were collected for each clinician, based on the same n=714 sample described for data mining above (as such only about half of the clinician respondents, n=33, had qualifying outcomes data). The principle questions were 1) how the survey results associated with the three components of TPB (Questions #1-9) would affect intentions (as indicated by Question #10), 2) if those effects would be mediated by other factors such as clinician experience or age, and 3) whether intentions were related to actual behavior (measured by adoption rates and outcomes). Survey and associated data were analyzed via correlation analysis, t-tests, ANOVAs, and regression analysis.

TABLE II. CLINICIAN PERCEPTION SURVEY

| Question # | TPB Component[a] | Question Text |
|---|---|---|
| CDOI 1 | PU | I can use CDOI to better understand my client's progress in therapy. |
| CDOI 2 | PU | I believe that CDOI is a good tool for building more effective relationships with clients. |
| CDOI 3 | PU | CDOI does not help me make decisions about what to do as a therapist. |
| CDOI 4 | NB | I am expected to use CDOI in my clinical setting at Centerstone. |
| CDOI 5 | NB | The therapists I work with, whose opinions I value support the use of CDOI. |
| CDOI 6 | NB | I believe that CDOI will become a permanent tool at Centerstone. |
| CDOI 7 | CB | CDOI does not fit my client population. |
| CDOI 8 | CB | My CDOI training prepared me to use the tool effectively. |
| CDOI 9 | CB | I have sufficient resources to use CDOI effectively (computers, space, privacy, time, etc.). |
| CDOI 10 | INT | I intend to consistently use CDOI to enhance my effectiveness as a therapist. |

a. PU = Perceived Utility, NB = Normative Beliefs, CB = Control Beliefs, INT = Intentions

## III. RESULTS

### A. Data Analysis

Descriptive statistics for ORS are presented in Table III, including baseline ORS, final ORS, and ORS final delta, broken out by state and by new vs. old clients. Additionally, to account for a regression to the mean effect, reliable change (where for ORS delta, <-4 equates to "deteriorate", between -4 and 4 equates to "no change", and >4 equates to "improve", as defined by Miller et al. (2006) was tested via Chi-Square for all clients (not excluding clients with baseline ORS>25) and found significant deviations from expectation ($\chi^2$=128.6, p<.000, n=714) given equal categorical expectations. It should be noted that approximately 52.8% (377/714) of clients achieved reliable improvement, which compares closely to Miller et al. [8] final values of 47%, although the reliable deterioration of 19.6% is much higher than their reported final 8%. Table IV shows the cross-tabulation breakout of reliable change and clinical significance (defined as moving from below clinical cutoff to above, for ORS cutoff=25, as per Miller et al. [8]) for clients who started in the clinical range (baseline ORS<=25), broken out by old and new clients. For the 184 new clients who fit this profile, nearly 46%, or approximately half, of those achieved both reliable improvement (reliable change = 3) and clinically significant change (clinical significance=1). Success was lower for old, existing clients, yet still 35.2% achieved both reliable improvement and clinically significant change.

TABLE III. DESCRIPTIVE STATISTICS

| State | | N | Min | Max | Mean | Std. Dev. |
|---|---|---|---|---|---|---|
| IN | bl_ors | 271 | 2 | 40 | 21.11 | 8.670 |
| | final_ors | 271 | 2 | 40 | 25.95 | 8.358 |
| | final_delta_ors | 271 | -18 | 30 | 4.85 | 8.412 |
| | Valid N | 271 | | | | |
| TN | bl_ors | 443 | 0 | 40 | 20.88 | 8.740 |
| | final_ors | 443 | 0 | 40 | 24.28 | 9.492 |
| | final_delta_ors | 443 | -25 | 38 | 3.40 | 9.435 |
| | Valid N | 443 | | | | |
| New[a] | | N | Min | Max | Mean | Std. Dev. |
| 0 | bl_ors | 461 | 0 | 40 | 21.43 | 8.811 |
| | final_ors | 461 | 0 | 40 | 24.68 | 9.342 |
| | final_delta_ors | 461 | -25 | 30 | 3.26 | 8.884 |
| | Valid N | 461 | | | | |
| 1 | bl_ors | 253 | 2 | 40 | 20.12 | 8.469 |
| | final_ors | 253 | 2 | 40 | 25.35 | 8.667 |
| | final_delta_ors | 253 | -18 | 38 | 5.20 | 9.318 |
| | Valid N | 253 | | | | |

a. 1=New, 0=Not New, see text for definition

TABLE IV. RELIABLE CHANGE VS. CLINICAL SIGNIFCANCE[a]

| Old Clients | | | Clinical Sign.[c] | | Total |
|---|---|---|---|---|---|
| | | | 0 | 1 | |
| Reliable Change[b] | 1 | Count | 46 | 0 | 46 |
| | | % of Total | 15.1% | .0% | 15.1% |
| | 2 | Count | 64 | 7 | 71 |
| | | % of Total | 21.1% | 2.3% | 23.4% |
| | 3 | Count | 80 | 107 | 187 |
| | | % of Total | 26.3% | 35.2% | 61.5% |
| Total | | Count | 190 | 114 | 304 |
| | | % of Total | 62.5% | 37.5% | 100.0% |

| New Clients | | | Clinical Sign.[c] | | Total |
|---|---|---|---|---|---|
| | | | 0 | 1 | |
| Reliable Change[b] | 1 | Count | 20 | 0 | 20 |
| | | % of Total | 10.9% | .0% | 10.9% |
| | 2 | Count | 36 | 1 | 37 |
| | | % of Total | 19.6% | .5% | 20.1% |
| | 3 | Count | 43 | 84 | 127 |
| | | % of Total | 23.4% | 45.7% | 69.0% |
| Total | | Count | 99 | 85 | 184 |
| | | % of Total | 53.8% | 46.2% | 100.0% |

a. Where Baseline ORS<=25
b. For Reliable Chance, 1=Deteriorate, 2=No Change, 3=Improve
c. For Clinical Significance, 1=yes (final ORS>25), 0=No (final ORS<=25)

Modeling results to predict ORS delta final can be seen based on all clients (Table V) and on new clients only (Table VI) with and without pre-discretization (CAIM) of the predictor variables, sorted by AUC. Both of these sets of results are based on models produced using a Naïve Bayes wrapper for feature selection (for reference, models of comparable performance were built without feature selection for all clients, while models consisting of new clients only did improve performance via feature selection). For both all clients and new only, the best performing models were generally either Ensemble methods or Bayesian methods (Naïve Bayes, AODE, LBR). For all clients, the best performing models produced around 70-71% accuracy and .76-.77 AUC. For new clients only, the best performing models produced around 75-77% accuracy and .82-.83 AUC. These results indicate that the CDOI has the ability to serve as the basis for predictive models that may anticipate the effect of clinical treatment in terms of change over time with reasonable accuracy.

TABLE V. MINING RESULTS (ALL CLIENTS)

| 10X Cross-Val (partitioned) | | | | | | |
|---|---|---|---|---|---|---|
| Model | Binning | Accuracy | AUC | TP rate | FP rate | H |
| Naïve Bayes | CAIM | 70.4% | 0.7718 | 71.5% | 30.8% | 0.2730 |
| Lazy Bayesian Rules | CAIM | 69.9% | 0.7716 | 70.5% | 30.8% | 0.2724 |
| Naïve Bayes | Bin Target | 70.6% | 0.7703 | 71.0% | 30.0% | 0.2678 |
| Ensemble | Bin Target | 70.7% | 0.7699 | 71.4% | 31.1% | 0.2658 |
| AODE | CAIM | 70.9% | 0.7697 | 72.4% | 30.9% | 0.2660 |
| Classif via Linear Reg | Bin Target | 70.7% | 0.7683 | 71.3% | 30.0% | 0.2667 |
| Log Regression | Bin Target | 69.7% | 0.7678 | 70.2% | 30.8% | 0.2658 |
| MP Neural Net | Bin Target | 69.5% | 0.7665 | 70.4% | 31.7% | 0.2682 |
| MP Neural Net | CAIM | 70.6% | 0.7659 | 71.3% | 31.3% | 0.2582 |
| Ensemble | CAIM | 70.7% | 0.7590 | 71.7% | 31.4% | 0.2582 |
| Bayes Net - TAN | CAIM | 69.0% | 0.7561 | 71.1% | 33.1% | 0.2430 |
| Log Regression | CAIM | 70.9% | 0.7555 | 71.4% | 29.8% | 0.2532 |
| Bayes Net - K2 | CAIM | 68.3% | 0.7517 | 68.9% | 32.4% | 0.2349 |
| Vote | CAIM | 68.5% | 0.7497 | 69.9% | 33.1% | 0.2354 |
| K-Nearest Neighbor | CAIM | 69.6% | 0.7451 | 72.2% | 33.1% | 0.2398 |
| Bayes Net - TAN | Bin Target | 69.3% | 0.7392 | 69.3% | 30.6% | 0.2180 |
| Vote | Bin Target | 66.7% | 0.7383 | 68.4% | 35.3% | 0.2140 |
| Bayes Net - K2 | Bin Target | 68.8% | 0.7314 | 68.5% | 30.9% | 0.2117 |
| Random Forest | CAIM | 68.2% | 0.7185 | 68.9% | 32.7% | 0.1956 |
| Random Forest | Bin Target | 65.3% | 0.6979 | 67.5% | 37.1% | 0.1661 |
| J48 Tree | CAIM | 69.2% | 0.6948 | 69.1% | 31.7% | 0.1992 |
| J48 Tree | Bin Target | 65.8% | 0.6669 | 65.9% | 34.2% | 0.3338 |
| K-Nearest Neighbor | Bin Target | 60.8% | 0.6371 | 63.1% | 41.7% | 0.0857 |

TABLE VI. MINING RESULTS (NEW CLIENTS ONLY)

| 10X Cross-Val (partitioned) | | | | | | |
|---|---|---|---|---|---|---|
| Model | Binning | Accuracy | AUC | TP rate | FP rate | H |
| Ensemble | CAIM | 76.7% | 0.8337 | 71.6% | 17.6% | 0.4030 |
| Ensemble | Bin Target | 76.3% | 0.8250 | 73.2% | 20.8% | 0.4130 |
| Naïve Bayes | CAIM | 75.5% | 0.8224 | 71.3% | 20.2% | 0.3929 |
| Naïve Bayes | Bin Target | 73.9% | 0.8174 | 71.8% | 24.3% | 0.3612 |
| AODE | CAIM | 72.7% | 0.8147 | 69.0% | 23.6% | 0.3708 |
| Lazy Bayesian Rules | CAIM | 73.1% | 0.8141 | 68.4% | 21.7% | 0.3761 |
| Vote | CAIM | 71.9% | 0.8098 | 68.3% | 24.4% | 0.3695 |
| MP Neural Net | CAIM | 72.7% | 0.8064 | 68.8% | 33.2% | 0.3626 |
| Log Regression | Bin Target | 72.7% | 0.8021 | 71.4% | 26.2% | 0.3494 |
| Classif via Linear Reg | Bin Target | 71.1% | 0.7941 | 67.3% | 27.3% | 0.3159 |
| K-Nearest Neighbor | CAIM | 73.1% | 0.7930 | 69.6% | 23.4% | 0.3233 |
| Vote | Bin Target | 72.7% | 0.7895 | 71.4% | 26.2% | 0.3374 |
| MP Neural Net | Bin Target | 73.9% | 0.7875 | 71.4% | 33.9% | 0.3204 |
| Bayes Net - K2 | CAIM | 70.4% | 0.7827 | 67.2% | 26.7% | 0.3156 |
| Log Regression | CAIM | 66.4% | 0.7791 | 63.6% | 31.1% | 0.3575 |
| Bayes Net - TAN | Bin Target | 70.8% | 0.7788 | 68.1% | 24.9% | 0.3253 |
| Bayes Net - K2 | Bin Target | 68.4% | 0.7724 | 66.4% | 30.0% | 0.3289 |
| J48 Tree | Bin Target | 73.1% | 0.7657 | 67.6% | 20.2% | 0.2985 |
| Bayes Net - TAN | CAIM | 70.0% | 0.7582 | 65.6% | 25.4% | 0.3384 |
| Random Forest | Bin Target | 69.6% | 0.7492 | 66.7% | 27.8% | 0.2748 |
| Random Forest | CAIM | 68.8% | 0.7411 | 64.6% | 26.8% | 0.2526 |
| K-Nearest Neighbor | Bin Target | 69.2% | 0.7191 | 67.6% | 29.6% | 0.2036 |
| J48 Tree | CAIM | 70.0% | 0.7140 | 65.0% | 24.1% | 0.2387 |

An additional question was whether early ORS/SRS ratings (including baseline scores) were predictive of ORS delta final, including relative to other potential clinical/demographic predictors such as age, gender, diagnosis, and treatment modality variation. The results can be seen in Table VII. Consistently, bl_ors and third_delta_ors appeared as the most significant variables – which given the lack of a "regression to the mean" effect, is notable. This finding agrees with the theory of Miller [8], who argues that early change from treatment (first 3 visits) predicts long term change (e.g. change after 3 months). Equivalent findings have been reported in other diseases such as lung cancer [45]. Additionally, bl_srs was identified as an important predictor whenever pre-discretization (CAIM) was applied. Payor_grp also appeared numerous times, with age, gender, diag_cat, and county occasionally appearing. Quantifying the actual importance of each variable relative to odds ratios, where possible, is particularly informative. In this regard, the third_delta_ors was by far the most critical variable, with an odds ratio of 11.37 ± 6.03 (higher values more likely to show greater final ORS delta). Bl_ors followed, with an odds ratio of 7.86 ± 3.58, lower values being more likely to show greater final ORS delta. Both bl_srs and gender had above-1 odds ratios, though the lower confidence interval fell near or below 1 (the significance threshold). In terms of payor_grp, it appears that commercial clients were more likely to obtain better final ORS delta values. It is possible that this result is related to socio-economic status, but this interpretation warrants caution and further study. There were also some differences in diag_cat, but they were very slight and of questionable importance. It should be noted that because all clients received the same treatment modality during the pilot (outpatient therapy), we were not able to truly evaluate the influence of treatment modality variation, and its effect remains to be seen.

TABLE VII. FEATURE SELECTION ODDS RATIOS

| Variable | Odds Ratio | For >Mean Improve |
|---|---|---|
| bl_ors | 7.86 ± 3.58 | * More Likely to be lower |
| third_delta_ors | 11.37 ± 6.03 | * More Likely to be higher |
| gender | 1.76 ± .72 | * More Likely to be Male |
| payor_grp | 1.75 ± .85 | * More Likely to be Commercial, Not Sign. |
| bl_srs | 1.74 ± .75 | * More likely to be lower, Not Sign. |
| diag_cat | 1.74 ± .79 | * Mood Disorders and Subst Abuse more likely to improve, Not Sign. |

### B. Implementation Analysis

Table VIII shows the correlations between various components of TPB, clinician age, and behavior (adoption rate), as well as clinical outcomes. Both normative beliefs (spearman's rho=.45, p=.000, n=66) and perceived utility (spearman's rho=.25, p=.042, n=66) significantly correlated with intentions. However, there was not a significant relationship between intentions and eventual behavior (spearman's rho= -.178). This result represents a previously identified problem in which beliefs/attitudes may correlate with intentions, but intentions do not necessarily correlate with behavior [46,47]. The most likely explanation is that organizational requirements to collect the CDOI by management affected behavior in ways that mitigated the impact of intent. However, one specific question (#4) that sought to address that issue found that older, more experienced clinicians were less likely to agree that they were expected to use the tool. Additional regression analysis found that TPB components explained 30.3% of the variance in intent (F=8.983, p=.000, df=62) but an insignificant amount of actual behavior (adoption).

TABLE VIII. TPB FACTOR CORRELATIONS

| | | PU | NB | CB | Intent | Adopt Rate | age | bl ors | bl srs | final delta ors |
|---|---|---|---|---|---|---|---|---|---|---|
| PU | Correlation | 1.000 | | | | | | | | |
| | Sig. (2-tailed) | . | | | | | | | | |
| | N | 66 | | | | | | | | |
| NB | Correlation | .080 | 1.000 | | | | | | | |
| | Sig. (2-tailed) | .523 | . | | | | | | | |
| | N | 66 | 66 | | | | | | | |
| CB | Correlation | -.176 | -.109 | 1.000 | | | | | | |
| | Sig. (2-tailed) | .158 | .383 | . | | | | | | |
| | N | 66 | 66 | 66 | | | | | | |
| Intent | Correlation | .251* | .450** | -.072 | 1.000 | | | | | |
| | Sig. (2-tailed) | .042 | .000 | .565 | . | | | | | |
| | N | 66 | 66 | 66 | 66 | | | | | |
| Adopt Rate | Correlation | .177 | .120 | -.147 | -.178 | 1.000 | | | | |
| | Sig. (2-tailed) | .172 | .356 | .260 | .169 | . | | | | |
| | N | 61 | 61 | 61 | 61 | 61 | | | | |
| age | Correlation | -.150 | -.102 | -.111 | -.261* | .411** | 1.000 | | | |
| | Sig. (2-tailed) | .252 | .439 | .400 | .044 | .002 | . | | | |
| | N | 60 | 60 | 60 | 60 | 55 | 60 | | | |
| bl ors | Correlation | .204 | -.209 | -.341 | -.226 | -.140 | -.337 | 1.000 | | |
| | Sig. (2-tailed) | .254 | .242 | .052 | .205 | .470 | .074 | . | | |
| | N | 33 | 33 | 33 | 33 | 29 | 29 | 33 | | |
| bl srs | Correlation | -.114 | .061 | .146 | .207 | -.277 | -.328 | -.003 | 1.000 | |
| | Sig. (2-tailed) | .526 | .736 | .418 | .247 | .146 | .082 | .986 | . | |
| | N | 33 | 33 | 33 | 33 | 29 | 29 | 33 | 33 | |
| final delta ors | Correlation | -.213 | -.255 | .052 | .118 | -.120 | .059 | -.144 | -.159 | 1.000 |
| | Sig. (2-tailed) | .235 | .151 | .776 | .514 | .534 | .761 | .424 | .378 | . |
| | N | 33 | 33 | 33 | 33 | 29 | 29 | 33 | 33 | 33 |

*. Correlation is significant at the 0.05 level (2-tailed)
**. Correlation is significant at the 0.01 level (2-tailed)
PU = Perceived Utility, NB = Normative Beliefs, CB = Control Beliefs

## IV. DISCUSSION

The CDOI, a patient-reported outcome measure, was implemented in a large, real-world behavioral healthcare setting in order to evaluate previous findings from smaller controlled studies, such as Miller [8]. PROs provide an alternative window into treatment effectiveness based on client perception and facilitate detection of problems/symptoms for which there is no discernible measure (e.g. pain) [6]. The principle focus of this paper is on evaluating the utility of PROs for predictive modeling of outcomes in a live clinical setting. The CDOI was found to be a relatively good predictor of change over time, in terms of predicting final outcome delta based on baseline and early change scores, as predicted by Miller [8] and other domains such as cancer [45]. Both measures of symptomology/functioning (ORS) and therapeutic alliance (SRS) contributed to this predictive capacity. Other variables, such as gender, also potentially affect this predictive capacity. Using new clients only, algorithms were constructed that could successfully predict above or below mean change approximately 75-77% of the time, well in excess of the 50% random chance. Analysis also showed that the outcome deltas over time were not simply a matter of regression to the mean.

Most importantly, the predictive capacity of the CDOI underscores the utility of direct client feedback measures such as PROs and their potential use as the basis for next generation clinical decision support (CDSS) tools and personalized treatment approaches. Although many CDSS tools have been constructed previously using traditional outcome measures, PROs open a new and uncharted window into understanding and predicting clinical outcomes and treatment effectiveness.

Implementation efforts were evaluated within the framework of the Theory of Planned Behavior (TPB) [21]. Significant relationships were found between normative beliefs, perceived utility, and clinician intent to use the CDOI. However, intention to use the CDOI did not necessarily translate into actual adoption behavior, a phenomenon previously reported elsewhere [46,47]. Adoption behavior appeared to be impacted by clinician experience and age, with older, more experienced clinicians more likely to use the CDOI despite lower levels of perceived utility and intent. Neither TPB components nor intent had any relation to actual clinical outcomes using the CDOI. This study provides an example of a framework for an integrated implementation evaluation in the context of a larger technology or outcome implementation. This sort of approach is a necessity for understanding technologies such as clinical decision support and outcomes measures such as PROs in real-world settings and the process of adoption [24,25]

The study presented here was a pilot of using PROs such as CDOI in a large, real-world mental health setting. Future directions include a larger evaluation of the CDOI in the context of clinical decision support for treatment recommendations in a real-world setting, and a comparison of the utility of the CDOI to more traditional measures of clinical outcome such PHQ9 in such settings [48]. Such an approach can produce individualized treatment predictions, as shown the examples in Figures 1 and 2 based on similar outcome measures, providing a clinician the probability of average or better treatment response across a number of modality combinations [13].

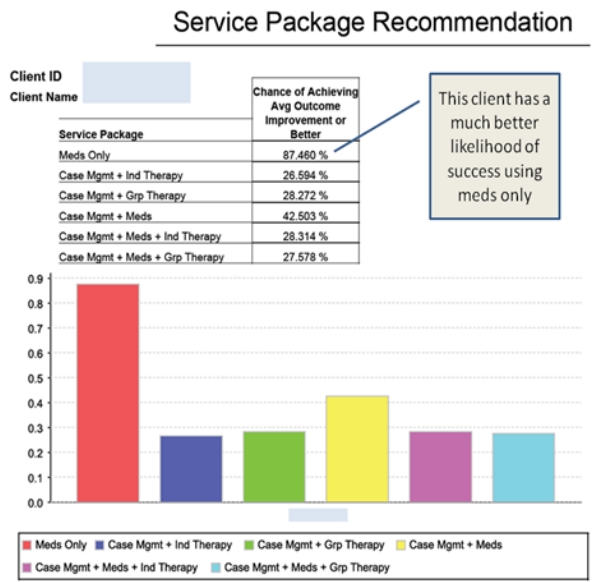

Fig. 1. Example 1 of treatment recommendations using pre-set "service packages" (from [12])

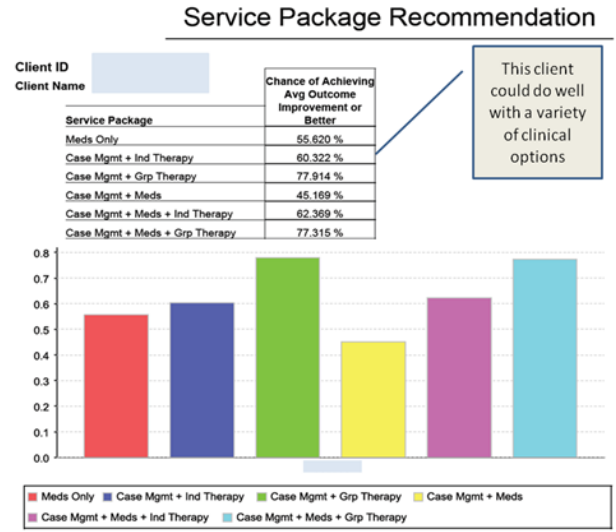

Fig. 2. Example 2 of treatment recommendations using pre-set "service packages" (from [12])

Utilizing PROs to enhance clinical decision-making through such data-driven CDSS approaches remains the long-term focus.

ACKNOWLEDGMENT

This research was funded by the Ayers Foundation and the Joe C. Davis Foundation. The opinions expressed herein do not necessarily reflect the views of Centerstone or its affiliates. The authors have no conflict of interest with the subjects described herein.

REFERENCES

[1] K. B. Ross, "Patient-reported outcome measures: Use in evaluation of treatment for aphasia," Journal of Medical Speech, vol. 14, no. 3, 2006.

[2] R. McCabe, M. Saidi, and S. Priebe, "Patient-reported outcomes in schizophrenia," The British Journal of Psychiatry. Supplement, vol. 50, pp. s21-28, Aug. 2007.

[3] L. S. Matza, K. S. Boye, and N. Yurgin, "Validation of two generic patient-reported outcome measures in patients with type 2 diabetes," Health and Quality of Life Outcomes, vol. 5, pp. 47-47, 2007.

[4] F. F. Duffy, H. Chung, M. Trivedi, D. S. Rae, D. A. Regier, and D. J. Katzelnick, "Systematic use of patient-rated depression severity monitoring: is it helpful and feasible in clinical psychiatry?," Psychiatric Services, vol. 59, no. 10, pp. 1148-1154, Oct. 2008.

[5] C. C. Gotay, C. T. Kawamoto, A. Bottomley, and F. Efficace, "The prognostic significance of patient-reported outcomes in cancer clinical trials," Journal of Clinical Oncology, vol. 26, no. 8, pp. 1355-1363, Mar. 2008.

[6] J. M. Valderas et al., "The impact of measuring patient-reported outcomes in clinical practice: a systematic review of the literature," Quality of Life Research, vol. 17, no. 2, pp. 179-193, Mar. 2008.

[7] J. M. Valderas, J. Alonso, and G. H. Guyatt, "Measuring patient-reported outcomes: moving from clinical trials into clinical practice," The Medical Journal of Australia, vol. 189, no. 2, pp. 93-94, Jul. 2008.

[8] S. D. Miller, B. L. Duncan, J. Brown, R. Sorrell, and M. B. Chalk, "Using formal client feedback to improve retention and outcome: Making ongoing, real-time assessment feasible," Journal of Brief Therapy, vol. 5, no. 1, pp. 5–22, 2006.

[9] D. L. Bringhurst, "The reliability and validity of the Outcome Rating Scale: A replication study of a brief clinical measure," Journal of Brief Therapy, vol. 5, no. 1, 2006.

[10] A. Campbell and S. Hemsley, "Outcome Rating Scale and Session Rating Scale in psychological practice: Clinical utility of ultra-brief measures," Clinical Psychologist, vol. 13, no. 1, pp. 1–9, 2009.

[11] B. L. Duncan, S. D. Miller, J. A. Sparks, D. A. Claud, L. R. Reynolds, J. Brown, and L. D. Johnson, "The Session Rating Scale: Preliminary Psychometric Properties of a 'Working' Alliance Measure," Journal of Brief Therapy, vol. 3, no. 1, pp. 3-12, 2003.

[12] M. G. Anker, B. L. Duncan, and J. A. Sparks, "Using client feedback to improve couple therapy outcomes: a randomized clinical trial in a naturalistic setting," Journal of Consulting and Clinical Psychology, vol. 77, no. 4, pp. 693-704, Aug. 2009.

[13] C. Bennett and T. Doub, "Data mining and electronic health records: Selecting optimal clinical treatments in practice," in Proceedings of The 6th International Conference on Data Mining, pp. 313-318, 2010. http://www.openminds.com/library/110410dmehr.htm

[14] C. C. Bennett, "Clinical productivity system: A decision support model." International Journal of Productivity and Performance Management, vol. 60, no. 3, 2011.

[15] I. S. Kohane, "The twin questions of personalized medicine: who are you and whom do you most resemble?," Genome Medicine, vol. 1, no. 1, pp. 4-4, 2009.

[16] O. Gevaert, F. De Smet, D. Timmerman, Y. Moreau, and B. De Moor, "Predicting the prognosis of breast cancer by integrating clinical and microarray data with Bayesian networks," Bioinformatics, vol. 22, no. 14, pp. e184-190, Jul. 2006.

[17] Y. Sun, S. Goodison, J. Li, L. Liu, and W. Farmerie, "Improved breast cancer prognosis through the combination of clinical and genetic markers," Bioinformatics, vol. 23, no. 1, p. 30, 2007.

[18] A. Boulesteix, C. Porzelius, and M. Daumer, "Microarray-based classification and clinical predictors: on combined classifiers and additional predictive value," Bioinformatics, vol. 24, no. 15, pp. 1698-1706, Aug. 2008.

[19] D. R. Hatfield and B. M. Ogles, "The use of outcome measures by psychologists in clinical practice.," Professional Psychology: Research and Practice, vol. 35, no. 5, p. 485, 2004.

[20] L. Bickman, "A measurement feedback system (MFS) is necessary to improve mental health outcomes," Journal of the American Academy of Child and Adolescent Psychiatry, vol. 47, no. 10, pp. 1114-1119, Oct. 2008.

[21] I. Ajzen, "The theory of planned behavior," Organizational Behavior and Human Decision Processes, vol. 50, no. 2, pp. 179-211, 1991.

[22] K. Kawamoto, C. A. Houlihan, E. A. Balas, and D. F. Lobach, "Improving clinical practice using clinical decision support systems: a systematic review of trials to identify features critical to success," British Medical Journal, vol. 330, no. 7494, pp. 765-765, 2005.

[23] J. A. Durlak and E. P. DuPre, "Implementation matters: a review of research on the influence of implementation on program outcomes and the factors affecting implementation," American Journal of Community Psychology, vol. 41, no. 3, pp. 327-350, Jun. 2008.

[24] D. F. Sittig et al., "Grand challenges in clinical decision support," Journal of Biomedical Informatics, vol. 41, no. 2, pp. 387-392, 2008.

[25] R. K. McHugh and D. H. Barlow, "The dissemination and implementation of evidence-based psychological treatments. A review of current efforts," The American Psychologist, vol. 65, no. 2, pp. 73-84, Mar. 2010.

[26] A. L. Francke, M. C. Smit, A. J. de Veer, and P. Mistiaen, "Factors influencing the implementation of clinical guidelines for health care professionals: A systematic meta-review," BMC Medical Informatics and Decision Making, vol. 8, pp. 38-38.

[27] M. Lugtenberg, J. M. Zegers-van Schaick, G. P. Westert, and J. S. Burgers, "Why don't physicians adhere to guideline recommendations in practice? An analysis of barriers among Dutch general practitioners," Implementation Science, vol. 4, pp. 54-54.

[28] M. R. Berthold et al., "KNIME: The Konstanz Information Miner," in Data Analysis, Machine Learning and Applications, 2008, pp. 319-326.

[29] D. J. Hand, H. Mannila, and P. Smyth, Principles of data mining. MIT Press, 2001.

[30] I. H. Witten and E. Frank, Data Mining: Practical Machine Learning Tools and Techniques, Second Edition, 2nd ed. Morgan Kaufmann, 2005.

[31] L. A. Kurgan and K. J. Cios, "CAIM Discretization Algorithm," IEEE Trans. on Knowl. and Data Eng., vol. 16, no. 2, pp. 145-153, 2004.

[32] H. Zhang, L. Jiang, and J. Su, "Hidden naive Bayes," in Proceedings of the 20th national conference on Artificial intelligence - Volume 2, pp. 919-924, 2005.

[33] G. I. Webb, J. R. Boughton, and Z. Wang, "Not so naive bayes: Aggregating one-dependence estimators," Machine Learning, vol. 58, no. 1, pp. 5–24, 2005.

[34] L. Breiman, "Random Forests," Machine Learning, vol. 45, pp. 5–32, Oct. 2001.

[35] J. R. Quinlan, C4. 5: programs for machine learning. Morgan Kaufmann, 2003.

[36] D. W. Aha, D. Kibler, and M. K. Albert, "Instance-Based Learning Algorithms," Machine Learning, vol. 6, pp. 37–66, Jan. 1991.

[37] R. Caruana, A. Niculescu-Mizil, G. Crew, and A. Ksikes, "Ensemble selection from libraries of models," in Proceedings of the twenty-first international conference on Machine learning, pp. 18, 2004.

[38] L. I. Kuncheva, Combining Pattern Classifiers: Methods and Algorithms. Wiley-Interscience, 2004.

[39] T. Fawcett, "ROC graphs: Notes and practical considerations for researchers," Machine Learning, vol. 31, 2004.

[40] D. Hand, "Measuring classifier performance: a coherent alternative to the area under the ROC curve," Machine Learning, vol. 77, no. 1, pp. 103-123, Oct. 2009.

[41] D. J. Hand, "Classifier Technology and the Illusion of Progress," Statistical Science, vol. 21, no. 1, pp. 1-14, Feb. 2006.

[42] A. Dupuy and R. M. Simon, "Critical review of published microarray studies for cancer outcome and guidelines on statistical analysis and reporting," Journal of the National Cancer Institute, vol. 99, no. 2, pp. 147-157, Jan. 2007.

[43] Y. Saeys, I. Inza, and P. Larrañaga, "A review of feature selection techniques in bioinformatics," Bioinformatics (Oxford, England), vol. 23, no. 19, pp. 2507-2517, Oct. 2007.

[44] R. Kohavi and G. H. John, "Wrappers for feature subset selection," Artificial Intelligence, vol. 97, pp. 273–324, 1997.

[45] D. T. Eton, D. L. Fairclough, D. Cella, S. E. Yount, P. Bonomi, and D. H. Johnson, "Early change in patient-reported health during lung cancer chemotherapy predicts clinical outcomes beyond those predicted by baseline report: results from Eastern Cooperative Oncology Group Study 5592," Journal of Clinical Oncology, vol. 21, no. 8, pp. 1536-1543, Apr. 2003.

[46] M. B. Perkins et al., "Applying theory-driven approaches to understanding and modifying clinicians' behavior: what do we know?," Psychiatric Services (Washington, D.C.), vol. 58, no. 3, pp. 342-348, Mar. 2007.

[47] R. J. Holden and B. Karsh, "The technology acceptance model: its past and its future in health care," Journal of Biomedical Informatics, vol. 43, no. 1, pp. 159-172, 2010.

[48] K. Kroenke et al., "Optimized Antidepressant Therapy and Pain Self-Management in Primary Care Patients with Depression and Musculoskeletal Pain: A Randomized Controlled Trial," JAMA : the journal of the American Medical Association, vol. 301, no. 20, pp. 2099-2110, May. 2009.